# Le terme et le concept : fondements d'une ontoterminologie

Christophe Roche
Equipe Condillac – Laboratoire Listic
Campus Scientifique
73 376 Le Bourget du Lac cedex
christophe.roche@univ-savoie.fr
http://ontology.univ-savoie.fr

**Résumé** : La terminologie connaît depuis plusieurs années un tournant linguistique important. On s'intéresse aujourd'hui davantage aux mots et à leur utilisation en discours qu'à connaître les choses qu'ils peuvent dénoter. Si effectivement l'approche wüstérienne et l'approche normative sont difficilement applicables *stricto sensu* et que la terminologie a tout intérêt à s'approprier le *signifié*, il n'en demeure pas moins que tous les mots n'ont pas le même statut et que la terminologie ne se réduit pas à une lexicographie technoscientifique. La société numérique pose de nouveaux besoins, réclame une opérationnalisation des terminologies et réactualise le primat du concept pour de nombreuses applications – il suffit de penser aux problèmes que soulève l'ingénierie collaborative –. L'appellation *ontoterminologie* traduit ce besoin de replacer le concept et sa dénomination au centre de la terminologie, tout en préservant sa dimension sociolinguistique par la prise en compte des termes d'usage à travers la langue de spécialité. Si une conceptualisation se dit bien en langue naturelle, elle se définit dans un langage formel selon des principes épistémologiques où l'ontologie occupe une place prépondérante.

**Plan**

1. Le tournant linguistique
2. Les besoins d'opérationnalisation
3. La terminologie : un ensemble de pratiques
4. L'ontoterminologie
5. Conclusion

## 1. Le tournant linguistique

Les mutations technologiques et économiques de ces dernières années impactent profondément nos structures sociétales. La notion de *communauté* est devenue centrale, rendant encore plus cruciaux les besoins de communication et de partage de l'information et, par conséquence, les besoins en terminologie et en normalisation.

Si la terminologie, et de façon plus générale les langues de spécialité, connaissent un intérêt grandissant, force est de constater que la doctrine wüsterienne est difficilement applicable et que l'approche prescriptive soulève de nombreux problèmes tant au niveau de la définition des normes que de leur mise en œuvre. Les critiques des principes fondateurs de la terminologie semblent justifiées, et en particulier le premier d'entre eux, en citant le manuel de terminologie de Felber disciple de Wüster : « *Il convient de se rappeler que tout travail terminologique devrait être fondé sur des notions et non sur des termes* ». L'approche onomasiologique serait apparemment inadaptée à la réalité d'une pratique considérée avant tout comme langagière. Tout cela conduit à dénier à la terminologie un statut de discipline autonome et milite pour la « ramener » sous la coupe des sciences du langage.

La terminologie connait donc depuis plusieurs années un tournant linguistique indéniable, la réduisant parfois à une lexicographie technoscientifique. Il existe plusieurs raisons à cela. Avant tout parce que la terminologie est mobilisée au sein de discours liés à une pratique et relève donc de la langue, certes de spécialité. L'étude de la terminologie se focalise alors sur les mots et leur utilisation en discours avec une attention toute particulière pour les textes – un mot isolé n'ayant pas de sens –. On s'intéresse plus aux expressions linguistiques qui dénotent les choses qu'à savoir ce que sont les choses. Aujourd'hui *être* c'est *être dit* et non plus *être pensé*. Une autre raison est la difficulté à cerner ce que peut être un concept et son rôle dans la détermination du sens du terme. La confusion entre *conceptualisation* et *classification* d'une part – il suffit de penser à l'approche prototypique –, et *sens* et *signification* d'autre part ; ont pour conséquence au pire le rejet du concept, au mieux sa réduction à un *signifié normé* ou à un réseau de mots liés par des relations linguistiques. Certains iront jusqu'à dire que Wordnet est une ontologie.

L'approche est séduisante. La langue est un système : les textes et les mots qu'ils contiennent constituent des données objectives sur lesquelles nous pouvons appliquer des méthodes scientifiques. La sémantique différentielle et la sémantique distributionnelle (étude des cooccurrences) en sont de beaux exemples et les résultats des plus intéressants. La terminologie relève *ipso facto* des sciences du langage.

Mais peut-on réduire la terminologie à une branche de la linguistique et oublier sa dimension conceptuelle ? Le fait qu'un terme puisse être mobilisé au sein de discours de façon similaire à un signe linguistique l'identifie-t-il pour autant à un tel signe réduisant du même coup le concept à un signifié ? Il est vrai qu'aujourd'hui la scène est davantage occupée par la dimension purement linguistique de la terminologie ; et lorsque l'on invite des experts d'un domaine c'est principalement pour qu'ils témoignent de cette dimension de leur activité et rarement de la conceptualisation et de la représentation des objets de leur domaine.

Cependant, on ne peut comprendre un discours (écrit ou oral) que dans la mesure où l'on partage une même culture. Ainsi, la compréhension de figures de rhétorique, telles que l'ellipse ou la métonymie fréquentes dans les documents scientifiques et techniques, nécessite que les locuteurs s'accordent sur un même *extralinguistique* qui par définition n'appartient pas à la langue. Cette culture commune, cet extralinguistique, ne constituerait-il pas le cœur même de la terminologie ?

Que la linguistique puisse être mobilisée pour l'étude de la terminologie, c'est une évidence. Ce qui ne veut pas dire que la terminologie relève des sciences du langage. Toute pratique scientifique met en œuvre non pas une langue, mais plusieurs systèmes de signes – nous ne pouvons pas penser sans signe nous rappelle Frege –. Mais qui dirait que la chimie, la thermodynamique, la mécanique quantique, la conception de systèmes d'information relèvent de la linguistique alors que leur pratique, par la compréhension des objets du monde et par la recherche d'une langue la moins ambigüe possible, pour ne pas dire normalisée, les rattache *ipso facto* à la terminologie ?

On identifie aujourd'hui trop souvent la terminologie à sa manifestation langagière – *verbalisation* d'une pratique à travers une *langue de spécialité* dont l'étude relève bien de la linguistique – en oubliant que la *conceptualisation* et la *représentation* des objets du monde sont des activités centrales, si ce n'est premières, de la terminologie. Activités qui font de la terminologie une discipline scientifique à part entière. Et si la terminologie met en jeu différents systèmes sémiotiques – conceptualisation et représentation nécessitent leurs propres langages – elle n'est pas uniquement une science des signes, mais aussi une science des choses. Le fait que la structure lexicale ne se superpose pas à la structure conceptuelle du domaine en est une illustration.

Enfin, l'existence de disciplines définies comme autant de spécialisations de la terminologie : terminologie textuelle, terminologie conceptuelle, terminologie cognitive, socioterminologie, ethnoterminologie, etc. n'est pas le fait d'irréductibles partisans d'une autonomie de la terminologie, mais traduit bien le fait qu'elle ne peut être réduite à une branche d'une discipline donnée.

## 2. Les besoins d'opérationnalisation

Entendons-nous bien. Notre objectif n'est pas ici de nier les apports de la linguistique. Le *terme d'usage*, ce terme mobilisé par une *langue de spécialité*, donne bien lieu à interprétation et la terminologie a tout intérêt à s'approprier le *signifié*. L'analyse des discours scientifiques, la compréhension de documents techniques le réclament. Nous souhaitons, dans le cadre de cette présentation d'ouverture à notre conférence, insister sur le fait que la terminologie ne se réduit pas à une analyse du discours scientifique et technique, à la recherche du sens des termes ou à une lexicographie de spécialité. La terminologie est une discipline scientifique dont le principal objet est de comprendre le monde et de trouver les mots « justes » pour en parler. La terminologie est une discipline autonome qui requiert pour son étude de puiser à l'épistémologie, la logique et la linguistique.

Il existe en effet des domaines scientifiques et techniques qui nécessitent une conceptualisation du monde et la création de dénominations univoques de ses constituants. C'est-à-dire « *un moyen d'expression qui permette à la fois de prévenir les erreurs d'interprétation et d'empêcher les fautes de raisonnement* » pour citer à nouveau Frege. Ces domaines reposent d'une part sur une compréhension consensuelle des choses et d'autre part sur leur représentation à des fins de manipulation. Ils donnent lieu à la réalisation d'applications qui s'appuient sur la définition d'une théorie – comprise ici comme une conceptualisation permettant d'appréhender les objets du monde – qui permet une certaine *objectivité*[1] dans la description et la manipulation de faits –. Les aspects descriptif et raisonnement de ces domaines priment sur les discours auxquels ils peuvent donner lieu.

Prenons pour exemple les applications de l'ingénierie collaborative qui connaissent avec l'ère informatique un essor considérable. La conception et la fabrication d'un produit, qu'il soit manufacturé ou non, repose de plus en plus sur la collaboration de communautés de pratique qui, bien que partageant une, ou partie, d'une même « réalité », peuvent différer tant au niveau de leur vision du monde que de la façon d'en parler. La solution ici ne réside dans une démarche de traduction des langues de spécialité, mais dans la définition d'un format d'échange, plus que d'une langue, reposant sur une conceptualisation et des dénominations consensuelles. Il est à ce propos important de souligner que les termes de cette interlangue seront d'autant plus acceptés que d'une part ils n'appartiennent à aucune des langues vernaculaires – il est plus facile de créer de nouveaux termes que d'imposer ceux d'une communauté –, et que d'autre part leur lecture permet de *comprendre* le système notionnel.

---

[1] La théorie est « objective » au sens où elle est acceptée et partagée par une communauté. Les descriptions et leurs manipulations le sont au regard de la théorie qui en contraint la forme et l'interprétation.

Pour être encore plus précis dans nos illustrations, la définition du modèle conceptuel d'un système d'information est un exemple typique d'une démarche terminologique « classique ». L'objectif est ici, avant tout, la définition d'une conceptualisation du domaine permettant de décrire les objets du monde qui puisse donner lieu à une représentation manipulable d'un point de vue computationnel (calculable par un ordinateur). C'est ensuite la recherche d'une dénomination univoque des termes dont la signification est le concept dénoté : l'approche est onomasiologique. Pour qui a assisté et participé à des réunions de conception de systèmes d'information, on peut être étonné – en fait la démarche est naturelle pour un scientifique – de ce souci constant de vouloir sortir de la langue naturelle et de ses ambiguïtés. Et systématiquement d'entendre : « Qu'est-ce que cela veut-dire ? Qu'est-ce que cela veut dire de façon précise ? Quel en est le sens exact ? ». Souci constant de s'extraire de tout discours pour se référer à un « socle » stable de connaissances défini dans un système formel dont la syntaxe et la sémantique sont clairement définies.

Il existe donc de nombreuses applications, en particulier les applications liées au traitement de l'information, qui réclament une opérationnalisation des terminologies. Des terminologies centrées sur la notion de concept qui soient *consensuelles*, *cohérentes*, *précises*, *partageables*, *réutilisables* et *calculables*. Autant de propriétés qui font que la terminologie ne peut relever du seul domaine de la linguistique.

Il nous semble important de rappeler que même lorsqu'on réduit la terminologie à une analyse du discours scientifique et technique, elle suppose une conceptualisation préalable du domaine[2]. Cette conceptualisation n'est pas du ressort de la linguistique. Sa définition relève d'une démarche épistémologique dans son appréhension des objets du monde et d'une démarche logique et computationnelle dans sa formalisation et sa représentation à des fins de manipulation. Elle met en œuvre un langage formel et un langage computationnel, dont les règles diffèrent de la langue naturelle. *Si une conceptualisation se dit bien en langue naturelle, elle se définit dans un langage formel selon des principes épistémologiques.*

## 3. La terminologie : un ensemble de pratiques

Le discours scientifique et technique « mélange » différents systèmes sémiotiques. La langue naturelle côtoie un langage symbolique et joue par rapport à ce dernier le rôle d'une métalangue – d'une glose – décrivant, expliquant, interprétant le langage symbolique. Ces systèmes sémiotiques ne répondent pas aux mêmes lois. Ils sont mobilisés par différentes pratiques qui ensemble constituent la terminologie proprement dite.

Ces différentes pratiques sont liées à la compréhension des objets du monde, à leur représentation à des fins de manipulation et aux discours auxquels ils peuvent donner lieu. Même si ces pratiques sont liées, il est important de les distinguer et d'étudier les rapports qu'elles peuvent entretenir.

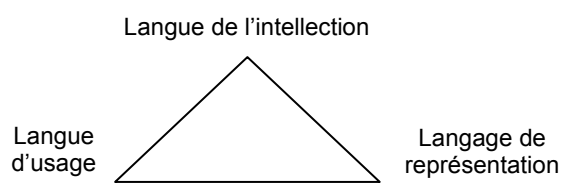

---

[2] Même dans le cadre d'une sémantique différentielle, la référence à cette conceptualisation est nécessaire à la compréhension des sèmes mais aussi à leur identification.

**a. La langue de spécialité (langue d'usage)**

Les discours[3] scientifiques et techniques relèvent de la *langue de spécialité*. Ils constituent, lorsqu'on étudie la terminologie d'un domaine sans en être un expert, la partie la plus visible et la plus directement accessible. Dans ce cadre, nous avons à faire à des *termes d'usage* qui donnent bien lieu à interprétation, à la recherche d'un sens qui se construit en discours. La notion de locuteur est centrale, et de façon plus générale la langue d'usage sous-entend la présence d'agents cognitifs tant au niveau de la production, et donc de l'intention, que de l'interprétation des discours. On s'intéresse ici aux rapports entre *signifiants* (termes d'usage) et *signifiés* en fonction d'un *contexte* donné. Cette pratique relève de la linguistique et de ses spécialités telles que la pragmatique.

L'analyse des discours, outre l'identification des termes d'usage, peut nous apporter une certaine connaissance du système notionnel[4]. Partant du fait que les documents scientifiques et techniques véhiculent des connaissances du domaine, il existe aujourd'hui de nombreux travaux qui portent sur l'extraction de connaissances, voire de terminologies, à partir de textes. L'existence de corpus numériques et l'utilisation de l'informatique permet d'obtenir des résultats intéressants (en particulier en sémantique distributionnelle). Cependant, il est indispensable de garder présent à l'esprit que l'incomplétude des textes est un des postulats de la linguistique textuelle. Ainsi, la compréhension des tropes suppose que l'auteur et le lecteur partagent un même *extralinguistique*. Mais comment prendre en compte l'intention de l'auteur à la base de toute interprétation, sachant qu'elle peut varier d'un texte à un autre au sein d'un même corpus ? *In fine* il est important de souligner que les structures lexicales et conceptuelles que l'on peut extraire de textes ne se superposent pas avec la conceptualisation du monde : *dire n'est pas concevoir*. L'oublier c'est aboutir à des systèmes non réutilisables dépendants d'un corpus donné qui ne peuvent être qualifiés de systèmes notionnels ni de terminologies. *La variabilité du signifié ne permet pas de cerner la stabilité du concept.*

Se focaliser uniquement sur le discours scientifique et technique, c'est oublier que la terminologie résulte avant tout d'une activité scientifique. C'est-à-dire d'une activité qui consiste à comprendre, modéliser et représenter un réel et des modes de raisonnement dans un système formel afin de décrire, vérifier et prédire certains faits. Cette activité, propre à l'ingénieur, suppose d'une part la capacité à appréhender la réalité et d'autre part la capacité à l'exprimer dans une théorie donnée. Pour cela *il est nécessaire de redonner à l'ingénieur une place centrale au sein de la terminologie*.

**b. La langue de l'intellection**

L'appréhension des objets du monde repose, en terminologie, sur le *concept*. Défini comme une « *unité de connaissance créée par une combinaison unique de caractères* » (norme ISO 1087), il permet de regrouper sous une même appellation les objets qui partagent des propriétés communes.

Un des mérites de la terminologie classique est d'avoir insisté à la fois sur l'importance d'une expression extralinguistique des concepts comme un ensemble de *caractères* et sur leur organisation en tant que système : « *Toute notion occupe une place définie dans un système particulier de notions* » (Manuel de terminologie. Felber). La détermination d'une typologie de caractères – caractères distinctifs, essentiels – et de relations entre concepts – logiques, ontologiques, de combinaison – relève d'une préoccupation principalement épistémologique sur la nature des connaissances descriptives indépendamment de leur expression dans une langue donnée, qu'elle soit naturelle (« *La partie qui traite des notions s'applique à n'importe quelle langue* » *ibidem*) ou formelle.

---

[3] oraux ou écrits (textes).
[4] Par exemple la recherche des relations définitoires, des relations d'hyperonymie et de méronymie considérées comme des expressions linguistiques des relations de subsomption et de mérologie. C'est aussi l'étude des adjectifs substantivants *versus* qualifiants, etc.

La combinaison et la factorisation de caractères n'est pas la seule façon de définir un concept. La recherche d'attributs donnés par l'expérience, l'identification de propriétés essentielles issues de la raison, la définition de fonctions à valeur prédicative sont autant d'approches qui correspondent à des principes épistémologiques – et à des choix idéologiques : empirisme, métaphysique, positivisme logique – qui guident mais aussi conditionnent la construction du système notionnel. *La terminologie dépend directement de la théorie du concept qui la fonde.*

**c. Les langages de représentation**

La représentation du système notionnel à l'aide d'un langage formel répond à deux besoins. Le premier correspond à une démarche scientifique. L'utilisation de langages symboliques à la syntaxe et la sémantique clairement définies permet de nous affranchir des problèmes d'interprétation que pose la langue naturelle. Accepter les axiomes et les règles d'un système formel, c'est en accepter les constructions et donc le système notionnel. Le deuxième besoin est un souci d'opérationnalisation. Le système notionnel doit pouvoir donner lieu à un modèle calculable par ordinateur.

Les différents formalismes de représentation ne nous assurent pas tous des mêmes propriétés. Celles de cohérence et de consensus sont certainement parmi les plus importantes. Elles conditionnent l'acceptation de la terminologie et par conséquence sa réelle utilisation.

Bien que les « principes terminologiques » (*ibidem*) soient d'inspiration logique, ils n'en ont pas toutes les qualités – les opérateurs sur les notions et les manipulations des différents caractères ne sont pas formellement définis –. Il ne serait pas aisé d'en définir un formalisme et un modèle computationnel satisfaisant. C'est la raison pour laquelle, lorsque l'on souhaite opérationnaliser une terminologie, on se tourne généralement vers d'autres systèmes.

La logique joue un rôle important dans la formalisation d'une conceptualisation. Elle est l'archétype des systèmes formels dont la syntaxe et la sémantique sont clairement définies. Le concept, fonction à valeur prédicative, est une *formule bien formée* et on dispose d'opérateurs et de mécanismes d'inférence pour la définition et l'exploitation des concepts. Enfin la logique est en elle-même un format d'échange. Autant de qualités qui nous garantissent certaines des propriétés recherchées dont la cohérence. A cela s'ajoute l'existence de logiques dédiées à la représentation des connaissances telles que les logiques des descriptions. *La logique est devenue aujourd'hui incontournable.*

Cependant les formalismes issus de l'intelligence artificielle demeurent les plus utilisés, principalement en raison de leur lisibilité : réseaux sémantiques, graphes conceptuels, systèmes à base de schémas. Le concept (ou classe) est défini par un ensemble d'attributs communs à ses instances. L'ensemble des concepts se structurent selon différentes relations : généralisation-spécialisation, partitive, etc.

**d. Le triangle sémiotique**

Distinguer les différentes pratiques – qu'elles relèvent du langage, de l'intellection ou de la représentation – qui toutes participent à la terminologie, c'est reconnaître à chacune son rôle et son importance sans vouloir en imposer une au détriment des autres. On peut ainsi reconnaître l'importance des termes d'usage (incluant variations terminologiques et figures de style) à côté des termes normés qu'il serait irréaliste de vouloir imposer. On peut également, par la séparation du concept et du signifié, permettre l'opérationnalisation des terminologies en garantissant un certain nombre de propriétés propres aux systèmes formels. En définitive, les notions mises en jeu – *signifiant, signifié, référent* versus *dénomination, concept, objet* – n'ont pas à être opposées à travers des triangles sémiotiques un peu réducteurs, mais ont tout à gagner à être mis en regard en insistant sur l'importance du contexte tant pour la définition de la conceptualisation (objectif, point de vue) que de la détermination du signifié (intention, interprétation).

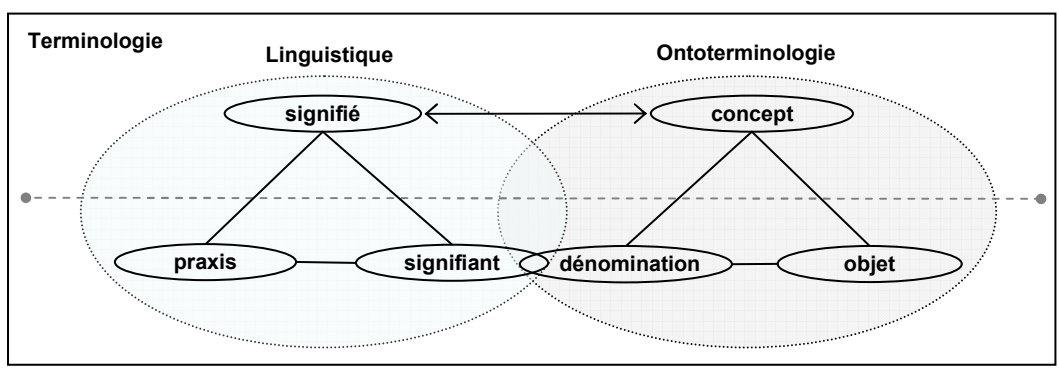

un *double* triangle sémiotique

# 4. L'ontoterminologie

Si l'utilisation d'un langage formel permet de s'abstraire des problèmes d'interprétation et d'ambiguïté que pose la langue d'usage et d'assurer certaines propriétés comme la cohérence et l'opérationnalisation des systèmes notionnels, elle ne permet pas de résoudre tous les problèmes et en particulier celui de la construction des systèmes notionnels. La logique en est un des exemples les plus significatifs. L'introduction de la *rigidité de prédicat*[5] (rigidité ontologique) illustre bien l'existence de connaissances de nature différente – différence entre propriété essentielle et propriété accidentelle – nécessaires à la compréhension d'une conceptualisation. Introduction, *a posteriori* et non *a priori*, d'une propriété qui conditionne les descriptions du monde mais ne guide en rien leur construction. S'accorder sur la syntaxe et la sémantique d'un langage formel n'est pas suffisant. Un système formel est avant tout un système de *réécriture* (de formules) et non une théorie de la connaissance ou une théorie linguistique[6]. Le problème central demeure celui de la construction du système notionnel et du choix des principes épistémologiques sur lesquels se reposer.

La terminologie classique insiste avec raison sur l'importance d'une définition extralinguistique du concept sous la forme d'une combinaison de caractères (propriétés ou qualités d'un objet) et sur l'importance des relations qui lient les concepts entre eux. Elle propose de plus un certain nombre de principes pour la construction du système notionnel basés sur la nature des connaissances en jeu et en particulier des caractères : intrinsèques - extrinsèques, restrictifs, etc. La *définition spécifique* d'une notion en est un exemple.

---
[5] Un prédicat est dit rigide si $\forall x [P(x) \rightarrow \Box P(x)]$ ; c'est-à-dire si P est vrai dans un monde possible, il l'est dans tous les mondes possibles. La rigidité relève *stricto sensu* d'une logique du second ordre (connaissance sur un prédicat).
[6] Les noms sont arbitraires et correspondent à des étiquettes sur des concepts.

Cependant il est à regretter une certaine confusion entre ce qui relève de préoccupations épistémologiques (classification des caractères, notions de genre et d'espèce, définition spécifique) et des systèmes formels (interprétation ensembliste des notions, opérateurs ensemblistes). Il en résulte trop d'imprécisions pour que les « principes terminologiques » puissent être directement utilisés : confusion entre notion et caractère dans leurs manipulations ; quelle est la définition intensionnelle de la notion résultante d'une disjonction ? Qu'en est-il des caractères distinctifs « hérités » par la conjonction (qui engendre une nouvelle espèce) de notions coordonnées créées par définition spécifique ? Et de façon plus générale, comment les opérateurs prennent-ils en compte la nature des caractères ? Comme si, sous l'influence d'un positivisme logique[7], on avait voulu faire passer ce qui relève de l'épistémologie – et donc d'une certaine façon de la métaphysique – sous les fourches caudines de la logique[8].

A cela s'ajoute un vocabulaire qui peut prêter à confusion sur l'emploi des mots *logique* et *ontologique* : la relation *genre-espèce* est qualifiée de « rapport logique » au même titre que l'intersection alors qu'elle relève de l'ontologie[9] ; les relations entre objets sont qualifiées de « rapports ontologiques » alors que leur mise en relation nécessite de les définir au préalable – pouvons-nous parler d'une chose sans la connaître ? –, c'est-à-dire de définir l'ontologie.

Les principes épistémologiques qui permettent d'appréhender les objets du monde et la construction du système notionnel constituent une problématique à part. Ils relèvent de l'ontologie proprement dite.

**a. Définition**

Nous introduisons le néologisme *ontoterminologie* pour désigner cette approche qui place l'ontologie au centre de la terminologie. Une approche où l'ontologie joue un rôle fondamental à double titre : pour la construction du système notionnel et pour l'opérationnalisation de la terminologie. L'*ontoterminologie* insiste d'une part sur l'importance des principes épistémologiques qui président à la conceptualisation du domaine – c'est l'ontologie dans sa définition première –, et d'autre part sur la nécessité d'une approche scientifique de la terminologie où l'ingénieur joue un rôle fondamental – c'est l'ontologie dans ses définitions plus récentes –. Ainsi, les représentations formelles de l'ontologie permettent de « sortir » de la langue naturelle et de garantir certaines propriétés comme la cohérence, le partage et parfois le consensus. Et ses représentations computationnelles autorisent une opérationnalisation des terminologies – les modèles calculables par ordinateur jouent pour la terminologie un rôle similaire à celui qu'a pu jouer et que joue la logique pour le langage en définissant un cadre de vérifiabilité des propositions théoriques –.

Regardons en quoi l'ontologie, dans ses différentes acceptions, constitue une aide précieuse pour la construction du système notionnel, et le cas échéant pour la création de mots « justes » pour en parler.

---

[7] Le Cercle de Vienne pour ne pas le nommer.
[8] Le système notionnel donne bien lieu à une interprétation ensembliste et à une interprétation logique à condition de les définir de manière formelle.
[9] L'ontologie et son interprétation logique (prédicat, syllogisme) sont deux choses différentes.

## b. Ontologie

L'ontologie[10], entendue comme « science de ce qui existe », constitue aujourd'hui une des voies les plus prometteuses pour la construction et la représentation formelle du système notionnel. C'est en particulier le cas pour la notion d'ontologie venant de l'intelligence artificielle, et plus précisément de l'ingénierie des connaissances. Issue de problèmes d'ingénierie collaborative au début des années 1990, elle vise des objectifs similaires à ceux de la terminologie classique : permettre la communication et l'échange d'information entre différentes communautés de pratique. Pour cela elle s'appuie sur une conceptualisation partagée d'un domaine sur laquelle repose la signification des termes. Les deux définitions suivantes résument la plupart des définitions existantes. La première insiste sur la dimension conceptuelle de l'ontologie : « *une ontologie est une conceptualisation d'un domaine – c'est-à-dire une définition formelle des concepts et de leurs relations – décrivant une réalité partagée par une communauté de pratique* » ; alors que la deuxième met en avant sa dimension terminologique – et normative – comme moyen de communication : « *une ontologie est un vocabulaire de termes dont les définitions sont données de manière formelle* ».
L'ontologie en ingénierie des connaissances est principalement un objet informatique, un moyen de représenter la réalité : *en intelligence artificielle, existe ce qui peut être représenté*. On comprend dès lors tout l'intérêt des ontologies pour la représentation du système notionnel et l'opérationnalisation de la terminologie. Mais le problème de leur construction reste entier.

L'ontologie est avant tout une théorie de la connaissance qui donne lieu à différents courants de pensée – et à différents principes épistémologiques – selon que l'on s'attache prioritairement à comprendre le monde où à le décrire tel qu'on le perçoit.

La définition des objets comme une somme de qualités perçues est une démarche naturelle et la plus immédiate. L'objectif ici n'est pas de comprendre le monde, mais de le décrire tel qu'il nous est donné, tel qu'on le perçoit à travers l'expérience, que ce soit par l'intermédiaire de nos sens ou de leurs prolongements technico scientifiques (appareils de mesure). Ces perceptions, que chacun partage parce qu'issues d'une expérience commune et sur lesquelles nous pouvons nous accorder (en particulier lorsqu'elles correspondent à des données scientifiques), définissent les qualités sur lesquelles se bâtit le système notionnel. Les concepts se construisent alors par abstraction, c'est-à-dire factorisation de qualités (caractères) communes : un concept est une « *unité de connaissance créée par une combinaison unique de caractères* » (ISO 1087-1), condition nécessaire et suffisante d'appartenance d'un objet à un concept. L'application itérative de ce processus d'abstraction aux différents ensembles de caractères permet de créer une structure notionnelle correspondant à un treillis[11] de concepts. Mais tout ensemble de caractères, s'il définit formellement un concept, n'est pas nécessairement porteur de sens. Cette démarche ne permet pas de prendre en compte les connaissances qui président à la formation et à l'organisation des concepts[12]. La factorisation de caractères reste une opération qui relève des systèmes formels.

---

[10] Bien que le mot lui-même soit de création récente (généralement attribué à Christian Wolff « Philosophia prima sive ontologica » 1729), l'ontologie est la « science de l'être ». Elle relève dans son acception première de la métaphysique et remonte aux origines de la philosophie.
[11] Ensemble (de concepts) muni d'une relation d'ordre partielle (inclusion sur les ensembles de caractères).
[12] Par contre cette approche, comme l'approche prototypique, semble bien adaptée à l'identification de concepts émergents dans un domaine en construction.

*Un concept est plus qu'une factorisation de qualités*. Les connaissances qui structurent les concepts en système relèvent de la raison et non de la perception. La démarche ici concerne l'ontologie dans son acception première de «*science de l'être en tant qu'être indépendamment de ses déterminations particulières*». On s'attache à comprendre ce que les choses sont, indépendamment de la façon dont elles peuvent être perçues. C'est-à-dire à rechercher les caractères *essentiels*[13] qui décrivent la nature de l'objet. Contrairement aux qualités – soumises « au plus et au moins » – qui décrivent l'état des objets, les caractères essentiels *définissent* et *différencient* les concepts. Issus de la raison et non de l'observation, ils structurent le système notionnel.

Le but est d'atteindre une description *stable* du monde sur laquelle on puisse s'accorder. Cette connaissance porte sur la *structure profonde* de la réalité. Souvent tacite[14], elle correspond à une *cristallisation* à un moment donné d'un savoir commun et partagé. Expliciter cette conceptualisation commune mais implicite est un problème difficile qui ne peut être résolu sans l'aide des experts du domaine. C'est la recherche de propriétés *objectives*, non pas de l'objet « en soi » indépendamment de tout observateur, mais de l'objet « pour soi » au regard d'une communauté de pratique. *L'ontologie est une modélisation d'une intersubjectivité.*

Quelle que soit la démarche – empirisme, métaphysique, logique – l'ontologie reste dépendante d'une pratique et non d'une langue qui découperait la réalité à la Sapir-Whorf. Elle n'est objective que dans la mesure où elle est partagée et acceptée par les membres d'une même communauté.

## c. Dénomination

Bien que la majorité des termes des domaines scientifiques et techniques soient motivés au sens où leur forme reflète dans une certaine mesure la structure du système notionnel (« relais de tension »[15] par exemple), il n'est pas toujours aisé de distinguer ceux qui relèvent du discours de ceux plus directement liés à la conceptualisation (« relais de tension » désigne-t-il un concept de même nom ou est-il uniquement un terme d'usage ?). En distinguant les termes d'usage des termes normés, on redonne au processus de dénomination[16] tout son intérêt. Le nom d'un concept n'est pas arbitraire : sa forme traduit (devrait traduire) la place du concept dans le système notionnel (« relais à seuil de tension » désigne le concept <relais à seuil de tension> subsumé par le concept <relais à seuil>). Le choix de la théorie du concept pour la construction du système notionnel impacte donc également la dénomination des concepts. Une théorie prenant en compte des caractères distinctifs ou essentiels apporte une aide indéniable – la définition des concepts par différenciation spécifique est l'exemple type : le nom de l'espèce se construit à partir de celui du genre (régissant) et de la différence (modificateur) –. *A contrario* comment nommer les concepts dans une approche purement logique où il n'existe que des fonctions à valeur de vérité ? Ces termes normés, s'ils n'ont pas à être imposés, sont indispensables à la désignation du système notionnel. Ils participent également à l'identification et à la définition des termes d'usage (le syntagme « relais de tension » est une expression d'usage de l'expression normée « relais à seuil de tension » avec ellipse du premier modificateur. Il ne désigne pas un concept différent).

---

[13] Un caractère est dit *essentiel* pour un objet si lorsqu'il est retranché de l'objet celui-ci n'est plus ce qu'il *est*.
[14] Cette conceptualisation tacite transparaît à travers certains termes d'usage dans l'emploi par exemple d'adjectifs substantivants – preuve supplémentaire s'il en était besoin de l'intérêt de l'analyse linguistique de corpus.
[15] Les unités lexicales sont notées entre guillemets et les concepts entre les symboles inférieur et supérieur.
[16] On préfèrera l'appellation *dénomination* pour insister sur une démarche onomasiologique à celle de *désignation* davantage liée à une utilisation du terme en langue.

### d. Exemple

La figure ci-dessous illustre une application de l'ontoterminologie à la gestion documentaire multilingue dans le domaine des échangeurs thermiques. Les documents sont indexés sur l'ontologie commune aux différentes communautés de pratique ; ce qui permet une recherche par concepts et non plus par mots clés. Ainsi, une recherche dans une langue donnée permet de retourner tous les documents, quelle que soit leur langue, associés aux concepts correspondant à la requête. Les termes d'usage et leurs relations linguistiques sont utilisés pour l'indexation automatique des documents et l'expansion (enrichissement) de requêtes. L'exploitation de l'ontologie, également utilisée pour l'expansion de requêtes (extension aux concepts subsumés), permet d'obtenir des résultats pertinents sans perte d'information. Enfin, l'ontologie permet un accès interactif à la base documentaire multilingue par les concepts du domaine. On parle également de cartographie sémantique ou de navigation sémantique interactive.

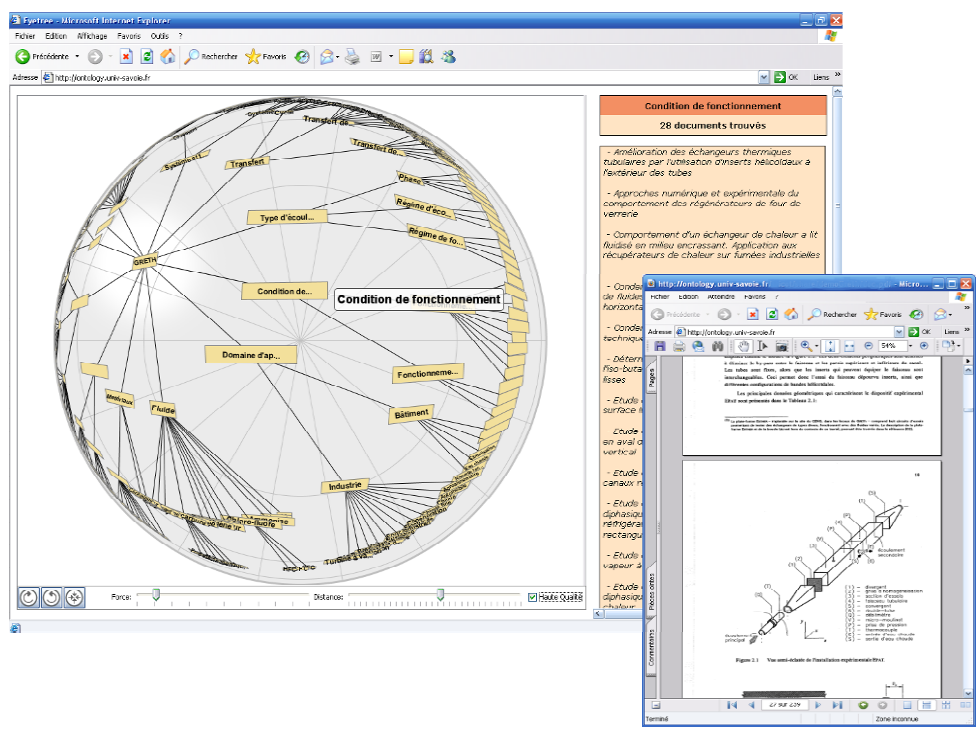

## 5. Conclusion

La terminologie ne peut se réduire à la seule étude des termes en langue. Ainsi, la détermination de leur sens requiert la connaissance préalable du système notionnel. De même, l'opérationnalisation des terminologies et la recherche de propriétés telle que la cohérence requièrent la mise en œuvre de systèmes formels détachés de tout discours. Le système notionnel, même s'il n'est pas toujours explicité, est au cœur de la démarche terminologique. Sa définition soulève de nombreux problèmes qui ne relèvent ni de la langue d'usage ni des formalismes de représentation.

La notion d'*ontoterminologie* met l'accent sur la dimension épistémologique de la terminologie dans son appréhension de la réalité. Elle permet de distinguer les pratiques – intellection, usage, représentation – et leurs fondements – terme d'usage *versus* dénomination, signifié *versus* concept –. Par la prise en compte de principes épistémologiques centrés sur la notion d'ontologie et de modèles computationnels respectant ces principes, l'ontoterminologie offre de nouvelles perspectives pour la construction de systèmes notionnels et leur représentation. Elle permet une construction du sens autour d'une sémantique référentielle et justifie l'intérêt de termes normés en regard des termes d'usage. En replaçant le concept et sa dénomination au centre de la terminologie, l'ontoterminologie redonne une place centrale à l'ingénieur dans son activité de conceptualisation. Enfin, elle propose des éléments de réponses aux enjeux de la société numérique et ouvre de nouveaux champs de recherche et d'applications.

## Bibliographie